# Characterizing Learning in Spiking Neural Networks with Astrocyte-Like Units

Christopher S. Yang, Sylvester J. Gates III, Dulara De Zoysa, Jaehoon Choe, Wolfgang Losert, and Corey B. Hart

*Abstract*—**Traditional artificial neural networks take inspiration from biological networks, using layers of neuron-like nodes to pass information for processing. More realistic models include spiking in the neural network, capturing the electrical characteristics more closely. However, a large proportion of brain cells are of the glial cell type, in particular astrocytes which have been suggested to play a role in performing computations. Here, we introduce a modified spiking neural network model with added astrocyte-like units in a neural network and asses their impact on learning. We implement the network as a liquid state machine and task the network with performing a chaotic time-series prediction task. We varied the number and ratio of neuron-like and astrocyte-like units in the network to examine the latter units' effect on learning. We show that the combination of neurons and astrocytes together, as opposed to neural- and astrocyte-only networks, are critical for driving learning. Interestingly, we found that the highest learning rate was achieved when the ratio between astrocyte-like and neuron-like units was roughly 2:1, mirroring some estimates of the ratio of biological astrocytes to neurons. Our results demonstrate that incorporating astrocyte-like units which represent information across longer timescales can alter the learning rates of neural networks, and the proportion of astrocytes to neurons should be tuned appropriately to a given task.**

## I. INTRODUCTION

THE design of artificial neural networks draws heavily from canonical theories of how neurons function. Biological neurons integrate information from upstream sensory stimuli and/or other neurons and convey this information to downstream neurons in the form of action potentials, i.e., spikes. Although there are many theories as to how information is encoded in spikes, one popular theory is that information is represented by the number of spikes per unit time, i.e., spike rate. Analogously, artificial neural networks are comprised of neuron-like units where the flow of information between units is controlled by weight and bias parameters (i.e., these parameters control the "spike rate" between units).

However, glial cells are another major type of cell which exist in the brain. Estimates of the ratio of glia to neurons range anywhere from 0.2:1 up to 10:1 [1, 2, 3, 4, 5]. This ratio is also variable and stratified based on region within the brain [2, 3, 4]. Traditionally, glial cells, of which there are many subtypes, were believed to simply support neurons. For instance, microglia are believed to mediate the brain's immune responses while oligodendrocytes ensheath neurons' axons with myelin, enabling faster conduction of signals through the brain. As a result, glia have largely been ascribed a non-computational function. However, more recent work has shown that astrocytes, one type of glial cell, play a crucial role in communication, learning, and processing within the brain [6]. Astrocytes possess extensions which can wrap around neuronal synapses and modulate neuronal communication. Astrocytes also communicate with other astrocytes via contact as well as the release of small molecules that modulate their activity.

Calcium activity in astrocytes is believed to play a particularly important role in brain computation. Neurons release neurotransmitters onto astrocytes, causing calcium levels in the astrocytes to increase which, in turn, cause the release of gliotransmitters back onto the neurons [7]. Calcium also flows between astrocytes in waves, enabling astrocytic modulation of synapses over large spatial scales. One key feature of calcium waves is that they propagate much more slowly than neuronal communication: while neurons communicate on the millisecond timescale, astrocytes communicate on the seconds timescale. This has led to suggestions that astrocytes are capable of representing information over longer temporal timescales than neurons.

Given the growing body of literature supporting a computational role for astrocytes, we investigated how incorporating separate neuron-like and astrocyte-like units into a neural network impacts the network's learning ability. To mimic the proposed functional role of astrocytes and neurons, we designed units which represent information over long and short timescales, respectively. We integrated these units into a liquid state machine (a type of reservoir computer) and trained the network to predict the chaotic dynamics of the Lorenz attractor. To examine the impact of the astrocytes on learning,

This work was supported by AFOSR award FA9550-21-1-0352. *Corresponding author: Christopher S. Yang.*

Christopher S. Yang is with Lockheed Martin Advanced Technology Laboratories, Cherry Hill, NJ 08002 USA (e-mail: christopher.s.yang@lmco.com).

Sylvester J. Gates III is with the Department of Biology, University of Maryland, College Park, MD 20742 USA (e-mail: sgates1@umd.edu)

Dulara De Zoysa is with the Department of Bioengineering, University of Maryland, College Park, MD 20742 USA (e-mail: dularadz@umd.edu)

Jaehoon Choe is with Lockheed Martin Advanced Technology Laboratories, Cherry Hill, NJ 08002 USA (e-mail: jaehoon.choe@lmco.com).

Wolfgang Losert is with the Department of Physics, University of Maryland, College Park, MD 20742 USA (e-mail: wlosert@umd.edu)

Corey Hart is with Lockheed Martin Aeronautics, King of Prussia, PA 19406 (e-mail: corey.hart@lmco.com)

we varied the ratio of neuron-like and astrocyte-like units across different training runs and compared learning between runs.

## II. Related Work

Reservoir computers are a type of recurrent neural network where units are randomly connected together in a reservoir with random fixed weights. The reservoir acts like a kernel function, nonlinearly mapping input signals into a high dimensional space [8]. A simple linear readout, such as a linear regression or a fully connected layer, is then used to extract information from the network. Reservoir computers are commonly used for time-series prediction tasks, particularly for chaotic time series [9]. One of the primary advantages of reservoir computers over other recurrent neural networks are that they are relatively simple to train. Whereas typical recurrent neural networks require training to tune all the parameters of the model, the reservoir's parameters remain fixed during training and only the linear readout's parameters are learned. Thus, reservoir computers require less training time as they generally learn fewer parameters.

The present study focuses on a specific type of reservoir computer known as a liquid state machine (LSM) where the units within the reservoir are a type of spiking unit, commonly leaky integrate-and-fire neurons. These units receive input in the form of spikes from upstream units, and when a unit's activity level crosses a threshold from receiving a sufficient number of spikes, it in turn generates a spike to send to downstream units. Thus, spikes are the basic computational currency of LSMs.

Previous studies have demonstrated that incorporating astrocyte-like units into reservoir computers can improve performance in machine learning tasks. For instance, Gergel' & Farkaš demonstrated that echo state networks (another type of reservoir computer) with astrocyte-like units outperform networks without astrocytes [10]. Abed and colleagues integrated astrocyte-like units into a simple spiking neural network and demonstrated that the astrocytes could increase the firing rate of neurons [11], although the network was not used to perform any machine learning tasks. Other studies have incorporated astrocyte-like units into neural networks for different purposes such as for repairing networks with faulty synapses [12].

There are different hypotheses as to why astrocyte-like units can improve the performance of these networks. One suggestion is that appropriately designed astrocyte-like units can drive a network to the edge of chaos. Performing computation near the phase transition between order and chaos can improve the computational capacity of the reservoir and thereby increase performance [13, 14]. Another suggestion is that a mismatch in the timescale of computing elements is more likely to optimally exploit their computational capacity [15].

The idea of representing information across different timescales for time-series prediction is not new. Long short-term memory (LSTM) networks do exactly this, representing short-term information as the "hidden state" and long-term information as the "cell state". While LSTMs perform well in general time-series prediction tasks, recent evidence suggests reservoir computers may be better suited for specifically predicting chaotic time series [9].

Whereas the machine learning community has generally focused on improving the performance of networks by incorporating *simplified* versions of astrocytes into their networks, neuroscientists have separately developed *biophysically detailed* models of astrocytes. These models span different levels of organization from modeling single astrocytes up to networks of astrocytes and neurons communicating with each other. For a more complete review of these biophysical models, see [16]. Here, we do not seek to implement a biophysically accurate version of an astrocyte. Rather, we seek inspiration from biological astrocyte function, designing units which incorporate elements of biological function.

## III. Methodology

### A. Neuron and astrocyte models

To build neuron-like and astrocyte-like units, we leveraged existing spiking unit models in snnTorch, a PyTorch-based library for training spiking neural networks. For simplicity, we model neurons as leaky integrate-and-fire (LIF) neurons. LIF neurons send information to other neurons in the form of spikes, which are generated when the neuron's membrane potential, $U$, crosses some threshold, $U_{thr}$. The membrane potential increases when a neuron receives an input spike from an upstream neuron, causing an influx of current, $I$. The potential also decays to zero over time where the decay rate is set by $\beta$. Upon generating a spike, the membrane potential instantaneously decreases by $U_{thr}$ to set the potential to a subthreshold state. The membrane potential dynamics are governed by the following equation:

$$U[t + 1] = \beta U[t] + I[t + 1] - R U_{thr}. \qquad (1)$$

Here, $R$ is a reset mechanism where $R = 1$ if $U[t + 1] > U_{thr}$ and $R = 0$ otherwise.

We implemented astrocyte-like units as a 2nd order LIF neuron in snnTorch, which have membrane potential dynamics as follows:

$$U[t + 1] = \beta U[t] + I_{syn}[t + 1], \qquad (2)$$

$$I_{syn}[t + 1] = \alpha I_{syn}[t] + I_{in}[t + 1]. \qquad (3)$$

Equation 2 is similar to Equation 1 in that $U$ decays to zero over time, but whereas in Equation 1 input spikes generate instantaneous deflections in membrane potential via $I$, in Equation 2 the synaptic current, $I_{syn}$, possesses its own dynamics that are governed by Equation 3. The synaptic current itself increases upon arrival of an input spike via $I_{in}$ and decays to zero over time proportional to $\alpha$. The addition of $I_{syn}$ enables a single input spike to not just generate an instantaneous increase in $U$ that subsequently decays but rather cause the membrane potential to rise over multiple timesteps according to the value of $\alpha$. Thus, the 2nd order unit mimics the long-term influence of astrocytes on downstream units. Additionally, we have modified the original 2nd order LIF neuron's behavior by removing the membrane potential reset mechanism as astrocytes do not possess such a reset mechanism. Thus, our astrocyte model exerts influence on downstream units while its membrane potential is above threshold and will continue to do so until the potential naturally decays to subthreshold levels.

### B. Network architecture

To form the reservoir, we generated four separate subnetworks of all-to-all recurrently connected units, two subnetworks of neurons and two of astrocytes (Figure 1A). We then connected these subnetworks together to mimic the transfer of information between neurons and astrocytes at the synapse, also known as the tripartite synapse. Each subnetwork of neurons possesses all-to-all downstream connections to one astrocyte subnetwork and as well as the other neuron subnetwork. Each astrocyte subnetwork receives all-to-all inputs from one neuron subnetwork and sends all-to-all outputs to the other neuron subnetwork.

Time-series data is input to the reservoir via a fully-connected layer to one of the neuron subnetworks. The entire time series is fed into the reservoir simultaneously where each datapoint is treated as a separate feature. After the data has been input to the reservoir $n$ number of times, a 3-layer multilayer perceptron reads out the activity of both neuron subnetworks and predicts the time series a predefined number of timesteps into the future (see IV.C for more information).

### C. Network training

We trained the network to predict the future state of the Lorenz system

$$\dot{x} = \sigma(y - x), \tag{4}$$

$$\dot{y} = x(\rho - z) - y, \tag{5}$$

$$\dot{z} = xy - \delta z, \tag{6}$$

where $x$, $y$, and $z$ are the state variables and $\sigma$, $\rho$, and $\delta$ are parameters that define the Lorenz system. We utilized a range of parameters for the Lorenz system generated probabilistically within predefined intervals. The $\sigma$ parameter was initialized at a value of 10 and added to values randomly from -5 and 5. Similarly, the $\rho$ parameter started at 28, subject to the same degree of random variation. The $\delta$ parameter was initialized at 2.667, with variations constrained to a smaller range of -0.5 to 0.5. We selected the Lorenz system because it exhibits chaotic dynamics, and reservoir computers have been previously shown to excel at predicting the trajectories of chaotic systems [17]. After simulating the system using Euler's method, we selected non-overlapping windows of 100 timesteps and divided each window in half, using the first 50 samples as input for prediction and the latter 50 samples as training/test/validation data to be predicted. The network was trained by minimizing the mean squared error between the predicted and actual trajectories using the Adam optimizer.

The data was divided into batches of size 32 and the prediction error with each presentation was preserved as a history of the network's loss function. We trained the network for 500 epochs and ran 100 iterations of the network. The number of neurons was incremented 10 times, using values drawn from the set $N = \{10, 50, 200, 300, 400\}$. The number of astrocyte-like cells was implemented as a proportion of each of these neuron counts. Proportions of astrocytes in the $m$th ($G_m$) run were calculated using the equation:

$$G_m = \left(\frac{3}{4} + \frac{n}{4}\right)N_m \quad n = 1, 2, \ldots, 10 \tag{7}$$

The network was run 10 times at each neuronal density, during which the proportion of astrocytes was varied, ranging over $G_m$ times the number of neurons at that iteration.

## IV. RESULTS

### A. Analysis of Learning Loss Function

Given the postulated astrocytic role in learning and memory we chose to test the impacts of including astrocyte-like effects at the synapses between connected neurons in our model. Two parameters immediately present themselves for exploration; the learning rate $\alpha$ (Figure 1B, orange arrow) and the loss after learning stabilizes (Figure 1B, black box region). The first parameter represents the rapidity with which the system metabolizes new information, while the second describes the absolute performance of the system.

We estimated the learning rate by calculating the slope of the curve over the first 10 consecutive epochs. The second parameter was estimated by excerpting the loss values starting at the point at which the average change in loss per epoch was less than 1% to the end and computing the average of this value. This was done for both training and validation loss curves. Learning rates were uniformly negative, as the system experienced little if any forgetting over these limited runs.

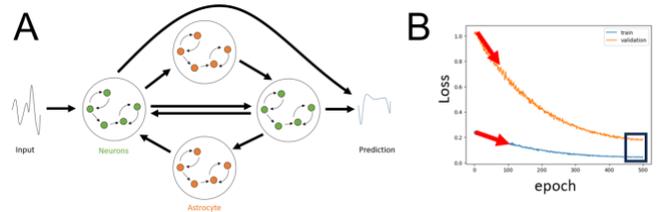

*Figure 1. A) Illustration of network architecture. Neuron-like (green) and astrocyte-like (orange) units were connected together to form the network's reservoir. The connectivity between the different types of units mimics the connectivity of the tripartite synapse between biological neurons and astrocytes. B) Analysis will measure the learning rate (slope of loss function, orange arrow) as well the end point loss value (black box) after some fixed number of training epochs during both training (blue) and validation (orange).*

We examined the role that network size played on the slope of the loss. We compared learning rate to both the aggregate size of the network including total elements (neuronal and astrocytic) as well as the individual concentrations of each modeled cell type. The learning rate grew more negative (a more rapid decrease in loss, essentially faster learning) with both overall network size and, broadly, the relative concentration of astrocyte-like elements in the data (marker size).

Similarly, the prediction error (averaged final loss value) for the network grew smaller with increasing number of each element individually as well their total.

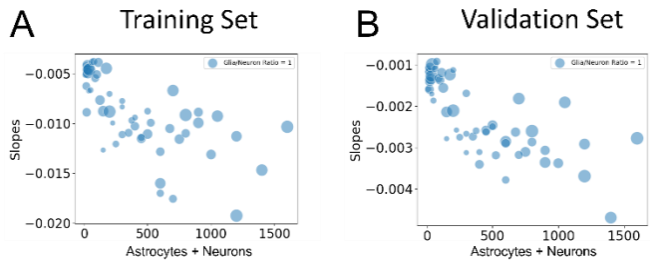

*Figure 2 A) The learning rate for training data increases (sharper decrease in learning loss) with increasing numbers of astrocytes in preparation. Small astrocyte to neuron ratios appear to be largely confined to smaller slope values. B) The same observations appear to hold true for the validation set.*

Given that that concentration of both modeled cell types, either individually or in concert, appeared to have an impact on learning performance, it is necessary to accurately assess the influence of the number of neurons and the number of astrocytes on the learning rate of our neural network. It is crucial to disentangle the individual contributions of these variables from the overall effect of the total number of elements (i.e., the sum of neurons and astrocytes). By employing multiple regression analysis, we can simultaneously evaluate the impact of each explanatory variable—neurons, astrocytes, and their combined sum—while controlling for the others. This statistical technique isolates the unique variance attributed to each variable, thereby providing a clearer understanding of how each contributes to the learning rate independently of the total number of elements. Including the sum of neurons and astrocytes as an additional variable allows for a more comprehensive analysis, facilitating more precise and nuanced inferences about the roles of neurons and astrocytes.

With those considerations in mind, we performed a multiple regression on those three variables. Interestingly, all three factors resulted in significant contributions to the algorithm's learning rate for both training. However, the analysis suggested significant collinearity between the factors, making the analysis less reliable. This situation can be handled by a LASSO regression [18] which can shrink redundant coefficients to zero, revealing the most important factors. With that in mind we performed LASSO on these factors and found that only the interaction term (A+N) made a significant contribution to the learning rate, LASSO zeroing out the individual neuronal and astrocytic density terms. The reconstruction using only the A+N term was quite good, with a correlation coefficient of 80% for the training data learning rate (Figure 3A) and 71% for the validation data learning rate (Figure 3B).

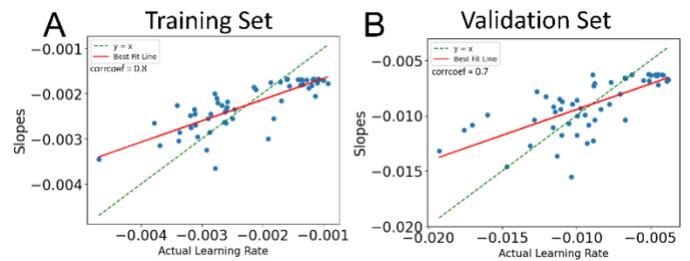

*Figure 3. A) Best fit LASSO model of learning loss slope from training epochs as a function of combined astrocytic and neuronal concentrations is very close to actual slope. B) Slightly weaker results for validation data.*

We performed the same LASSO analysis on the astrocytic, neuronal, and A+N factors for the averaged final loss value and found similar results (the A+N variable remained the most significant contributor to the loss value) but the correlation of the LASSO estimate with the original value was significantly less for both data sets (62% and 50% respectively). This is consistent with the observation that the effect of network size on endpoint loss to be relatively weak.

Given these observations, an obvious question presents itself: is there an optimal balance between the neuron count and the astrocyte count in this model, with respect to the impact on the model learning rate? We calculated the ratio of astrocyte counts to neuron counts (A/N) for all pairs of values in our simulations and constructed density plots of the learning rate data as a function of A/N ratios using a gaussian kernel (Figure 4). While the observed probability density functions are quite wide, the most negative learning rates clearly cluster at A/N ratios in the vicinity of 2:1, a number that is close in scale to at least some estimates of cortical ratios [1, 2, 3, 4, 5].

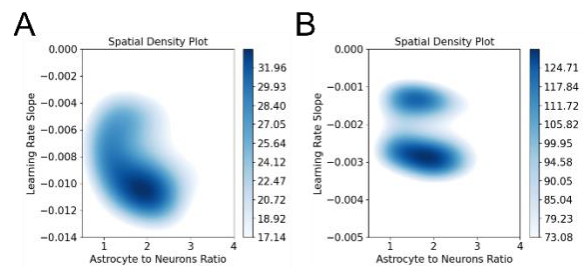

*Figure 4. Kernel density estimate of Slope/Learning Rate as a function of astrocyte to neuron ratio. Note both training and validation sets show strong concentration at almost 2:1 ratio.*

## V. DISCUSSION

It has become increasingly clear that the 'neuron doctrine' [19, 20], which underpins much of contemporary computational neuroscience, is not only incomplete but may also be fundamentally flawed. To explore these findings, we conducted a series of simulations to assess the impact of astrocyte-like elements on neural information processing.

Although the effects observed are modest, they are nonetheless significant, demonstrating that astrocyte-like information processing impacts learning in neural networks, and the proportion of neurons to astrocytes must be tuned appropriately for a given task.

Our results represent a preliminary attempt to clarify the information processing role of astrocytes in the nervous system. Of particular interest is the finding of a low A/N ratio which is close to some estimates within the literature. Future research could focus on leveraging these insights to advance machine learning methodologies. This can include looking into further methods to model astrocyte-like units and other glial cell types based on their biological impacts on neurons and brain activity.

### Acknowledgment

We thank members of the Losert lab for helpful discussions.

**Christopher S. Yang** received a B.A. in neuroscience from the University of Virginia (Charlottesville, VA, USA) in 2016 and a Ph.D. in neuroscience from Johns Hopkins University (Baltimore, MD, USA) in 2022. He is currently a senior research scientist at Lockheed Martin Advanced Technology Laboratories in Cherry Hill, NJ, USA. His current research interests include human-machine teaming, cognitive modeling, and neuromorphic computing.

**Sylvester J. Gates III** received a B.S. degree in biological sciences from the University of Maryland College Park (College Park, MD, USA) in 2015. He is currently a biological sciences Ph.D. candidate at the University of Maryland, College Park (College Park, MD, USA) with a concentration in cellular and molecular biology. His current research interests include stem cell models of human brain cells, live-cell dynamics and fluorescent imaging, and calcium dynamics.

**Dulara De Zoysa** received a B.S. in Biomedical Engineering (Gold medal for best academic performance) from University of Moratuwa (Moratuwa, Sri Lanka) in 2018. He is currently a PhD candidate in Bioengineering at the University of Maryland, College Park in College Park, MD, USA. He is also a member of the Phi Kappa Phi Honor Society. His current research interests include real-time analysis and optogenetic stimulation of neuronal cells, and collective learning of neuronal circuits.

**Jaehoon Choe** received a B.A. in biological sciences from the University of Chicago (Chicago, IL, USA) in 2004 and a Ph.D. in neuroscience from University of California, Los Angeles (Los Angeles, CA, USA) in 2014. He is currently a staff research engineer at Lockheed Martin Advanced Technology Laboratories in Arlington, VA, USA. His current research interests include human-machine teaming, neuro-symbolic systems, and machine-learning assisted assurance systems.

**Wolfgang Losert** received a Diplom in Technical Physics from the Technical University of Munich (Munich, Germany) in 1995 and a PhD in physics from the City College of New York (New York, NY, USA) in 1998. He serves as MPower professor of physics at the University of Maryland, College Park in College Park, MD, USA. Prof. Losert is a fellow of AAAS and the American Physical Society.

**Corey B. Hart** received a B.S. in physics from the Villanova University (Villanova, PA, USA) in 1995 and received his Ph.D. in neuroscience from the Pennsylvania State University (State College, PA, USA) in 2001. He is currently a principal research scientist at Lockheed Martin Advanced Development Programs (Fort Worth, TX, USA). His current research interests neuromorphic computing, advanced machine learning and non-Von Neumann computing architectures.